\documentclass{bmcart}

\usepackage[utf8]{inputenc} %unicode support
\usepackage{graphicx}
\usepackage{bm}
\usepackage{color}
\usepackage{subfigure}
\usepackage{multirow}
\usepackage{algorithm}
\usepackage{algorithmic}
\usepackage{array}
\usepackage{subfigure}
\usepackage{mathrsfs}
\usepackage{epstopdf}
\usepackage{amsmath}
\usepackage{enumerate}
\usepackage{amsmath}
\usepackage{framed}
\usepackage{footmisc}
\usepackage{endnotes}
\let\footnote=\endnote

\usepackage{booktabs,caption,fixltx2e}
\usepackage[flushleft]{threeparttable}
\definecolor{shadecolor}{rgb}{255,255,0} %背景色亮黄色
\startlocaldefs
\endlocaldefs

\begin{document}
%%% Start of article front matter
\begin{frontmatter}

\begin{fmbox}
\dochead{Research}

\title{Mining Functional Modules by Multiview-NMF of Phenome-Genome Association}

\author[
   addressref={aff1,aff2},                   % id's of addresses, e.g. {aff1,aff2}
   %corref={aff1},                       % id of corresponding address, if any
   %noteref={n1},                        % id's of article notes, if any
   email={ygzhang@mail.nankai.edu.cn}   % email address
]{\inits{YGZ}\fnm{YaoGong} \snm{Zhang}}
\author[
   addressref={aff1,aff2},
   email={xuyingjie@mail.nankai.edu.cn}
]{\inits{YJX}\fnm{YingJie} \snm{Xu}}
\author[
   addressref={aff1,aff2},
   email={nkufanxin@mail.nankai.edu.cn}
]{\inits{XF}\fnm{Xin} \snm{Fan}}
\author[
   addressref={aff1,aff2},
   email={hongyuxiang@mail.nankai.edu.cn}
]{\inits{YXH}\fnm{YuXiang} \snm{Hong}}
\author[
   addressref={aff1,aff2},
   email={liujiahui@mail.nankai.edu.cn}
]{\inits{JHL}\fnm{JiaHui} \snm{Liu}}
\author[
   addressref={aff1,aff2},
   email={hezhicheng@mail.nankai.edu.cn}
]{\inits{ZCH}\fnm{ZhiCheng} \snm{He}}
\author[
   addressref={aff1,aff2},
   email={huangyl@nankai.edu.cn}
]{\inits{YLH}\fnm{YaLou} \snm{Huang}}
\author[
   addressref={aff1,aff2},
   corref={aff1},                       % id of corresponding address, if any
   email={xiemq@nankai.edu.cn}
]{\inits{MQX}\fnm{MaoQiang} \snm{Xie}}

\address[id=aff1]{%
  \orgname{College of Software, Nankai University},
  %\street{D\"{u}sternbrooker Weg 20},
  \postcode{300350}
  \city{TianJin},
  \cny{China}
}
\address[id=aff2]{%                           % unique id
  \orgname{College of Computer and Control Engineering, Nankai University}, % university, etc
  %\street{WeiJin Road},                     %
  \postcode{300350}                                % post or zip code
  \city{TianJin},                              % city
  \cny{China}                                    % country
}

%\begin{artnotes}
%%\note{Sample of title note}     % note to the article
%\note[id=n1]{Equal contributor} % note, connected to author
%\end{artnotes}

\end{fmbox}% comment this for two column layout

\begin{abstractbox}

\begin{abstract} % abstract
%\begin{shaded}
\parttitle{Background}
Mining gene modules from genomic data is an important step to detect new gene members of the pathways or other relations such as protein-protein interactions. In this work, we explore the feasibility of detecting gene modules by factorizing gene-phenotype associations with phenotype ontologies rather than the conventionally used gene expression data. In particular, the hierarchical structure of the ontologies has not been taken full advantage of in clustering genes
and the consistency proposed is believed to be found in the gene clusters obtained with the method built on the hierarchical structure of ontologies.
%and the consistency mentioned in this paper is believed to be found in the gene clusters obtained according to hierarchical structure of the ontologies.
%while functionally related genes are consistently associated with phenotypes on the same path in the phenotype ontology.\\

\textbf{Results:}
We propose a hierarchal Nonnegative Matrix Factorization (NMF)-based method, called Consistent Multiple Nonnegative Matrix Factorization (CMNMF),
with which the genome-phenome association matrix has been factorized into two levels of hierarchical structure among phenotype ontologies so as to mine gene functional modules.
%CMNMF constrains the gene clusters from the association matrices at two consecutive levels to be consistent since the genes are annotated with both the child phenotypes and the parent phenotypes in the consecutive levels.
Gene clusters from the association matrices at two consecutive levels are constrained by CMNMF and are consistent since the genes are annotated with both the child phenotypes and the parent phenotypes.
CMNMF also restricts the identified phenotype clusters to be intensively connected within the phenotype ontology hierarchy. In the experiments on mining functionally related genes from mouse phenotype ontologies and human phenotype ontologies, CMNMF effectively improves clustering performance over the baseline methods. Gene ontology enrichment analysis is also conducted to reveal biologically significant gene modules. \\
\textbf{Conclusions:}
%Our work show that the proposed CMNMF can identify functional gene modules with biological significance over conventional methods. The performance of clustering results show the effectiveness of CMNMF. Our work provides a new perspective to mine functional gene modules and disease genes.\\
Utilizing the information in the hierarchical structure of phenotype ontologies, biologically significant gene modules can be identified with CMNMF.
CMNMF also serves as a better tool for detecting new gene members in the pathways and protein-protein interactions.\\
%\end{shaded}
\textbf{Availability:} \texttt{https://github.com/nkiip/CMNMF}\\
%\textbf{Contact:} {name@bio.com}{name@bio.com}\\
%\parttitle{Second part title} %if any
%Text for this section.
\end{abstract}


\end{abstractbox}
%
%\end{fmbox}% uncomment this for twcolumn layout

\end{frontmatter}

\section*{Background}
%\begin{shaded}
Gene functional modules within the genomic data are often identified to find genes sharing the same functions, being involved in the same pathway or interacting with each other.
To find these functionally related gene sets, clustering methods are commonly used. K-means and AHC (Agglomerative Hierarchical Clustering) are the most frequently used clustering algorithms to cluster gene expression data. Recently, NMF and its variants have also been successfully adopted for clustering gene expression \cite{Wang2013} and gene-phenotype association data \cite{Hwang2012}. NMF has advantages over other methods because of its interpretability and good performance.
%It is interesting to cluster genes by factorizing genome-phenome association data that will provide us a higher level to globally explore the relationship between genes and diseases.

More recently, multi-view NMF methods have been proposed. Collective NMF (ColNMF), proposed by Singh, can get more consistent results by using the shared coefficient matrix but different basis matrices across different views \cite{Singh2008}. Zhang and Zhou proposed a multiple NMF framework to integrate multiple types of genomic data to identify microRNA-gene regulatory modules jointly \cite{Zhang2011}. Additionally, NMF-based methods integrated with some structure information were also proposed and have achieved better results. Pehkonen adopted NMF to analyze association data between gene and Gene Ontology (GO), in which the association matrix has been enriched according to the ``true path rule" \cite{Valentini2010} of gene ontology hierarchy \cite{Pehkonen2005}. Hwang and Kuang proposed a nonnegative matrix tri-factorization method to cluster phenotypes and genes simultaneously \cite{Hwang2012}. Focusing on predicting missing traits for plants, HPMF incorporates hierarchical phylogenetic information into matrix factorization \cite{Shan2012}.

In the context of gene clustering, one useful structural feature that has never been integrated with multi-view NMF is the hierarchical structure phenotype ontology which is potentially helpful for clustering genes. In this work, we assume that the clustering of genes with different representations in multiple views should be consistent, i.e. genes will be consistently clustered by the association with phenotypes at different levels (or granularity) of the hierarchy in a phenotype ontology, in which each level of the phenotype ontology provides a different view for clustering genes.

Based on this motivation, we propose a multi-view NMF-based method called CMNMF (consistent multiple non-negative matrix factorization) for mining functional gene modules, in which the hierarchical structure of phenotype is introduced as prior knowledge.
In detail, a consistency constraint on gene clusters and a hierarchical mapping constraint among the phenotypes in two consecutive levels in the ontology are introduced in the loss function. In the experiment, we apply CMNMF on gene-phenotype association data of mouse and human. CMNMF is compared with the baseline methods by measuring the performance of predicting KEGG pathways and protein-protein interaction networks. Furthermore, GO enrichment analysis equipped with DAVID tool \cite{Dennis2003} is performed to evaluate the biological significance of the gene clusters.
%Finally, a supervised version of CMNMF is also proposed for predicting disease causal genes.
%\end{shaded}

\begin{figure}[!t]
  \centering
  \begin{minipage}{.80\linewidth}
  \centering
    \includegraphics[width=\linewidth]{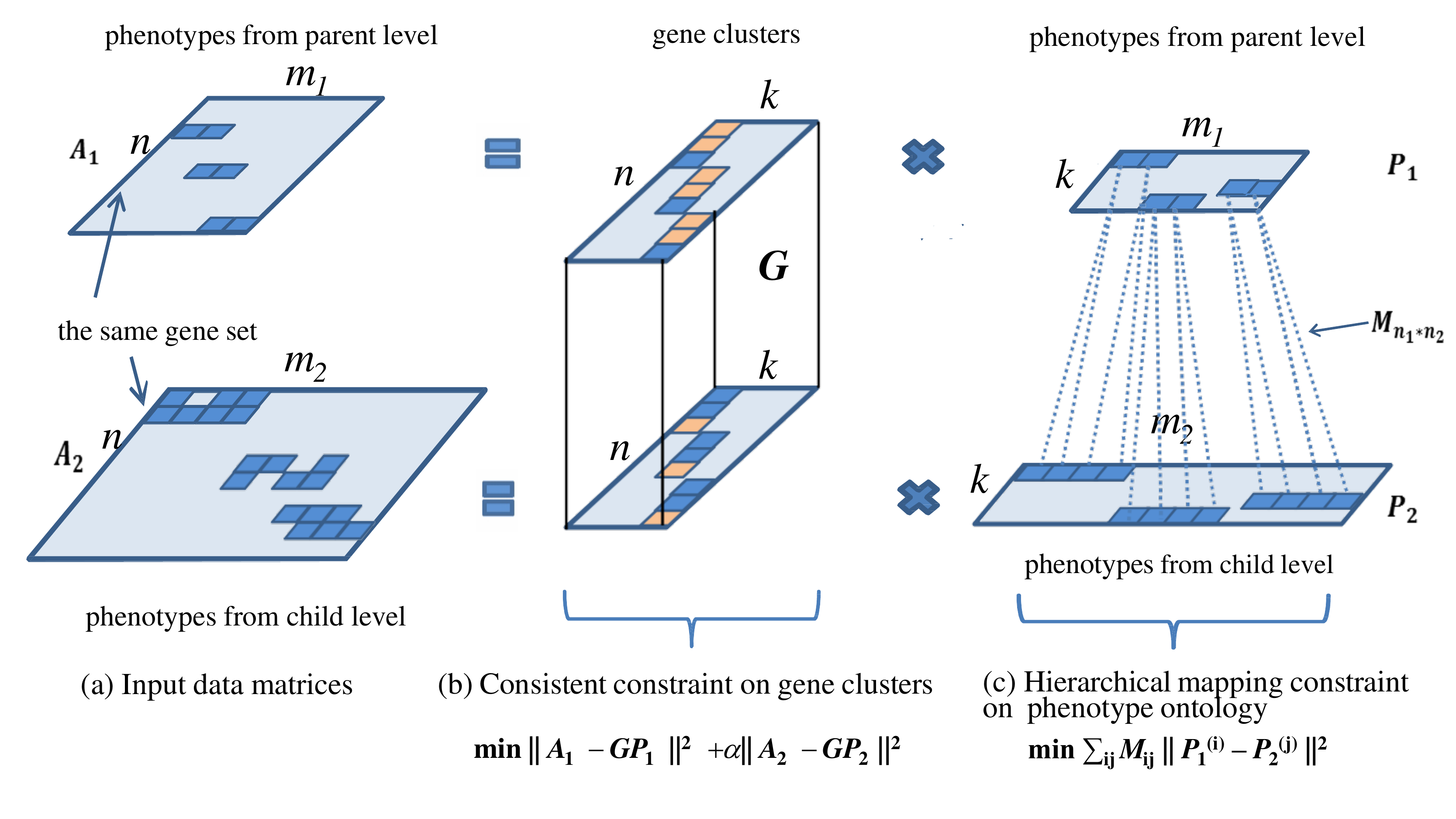}
  \end{minipage}
  \caption{Illustration of the CMNMF framework. (a) The gene-phenotype\\  associations are divided into two matrices according to the level of phenotype \\ ontologies, and the two matrices share the same gene set. (b) consistent const-\\raint on factorized gene clusters. (c) Hierarchical mapping constraint on the \\phenotype ontologies at parent and child levels.}
  \label{fig:model}
\end{figure}

\section*{Materials and Method}
\definecolor{shadecolor}{rgb}{255,255,0} %背景色亮黄色
\subsection*{\textbf{Data Preparation}}
Mouse gene-phenotype associations were downloaded from Mouse Genome Informatics (MGI)\cite{Blake2003} in Feb. 2016, including 15,524 associations between 5,971 phenotypes and 1,350 genes. More specifically, 3,414 phenotype ontologies were at level 7 and their parents, 2,557 phenotype ontologies were at level 6.
The phenotype ontology levels in our experiment were chosen by the ``most frequent level of annotation" criterion \cite{Espinosa2011}.
%these data have been applied in our experiments followed by the description in \cite{Espinosa2011}.
Two versions of mouse protein-protein interaction (PPI) network were obtained from the BIOGRID (Feb. 2016 and Sep. 2016) \cite{Chatr-Aryamontri2015} and two versions of 292 mouse KEGG pathways (Feb. 2016 and Sep. 2016) were extracted for evaluation \cite{Kanehisa2016}.

53,929 human gene-phenotype associations between 3,280 genes and 5,948 phenotypes were downloaded in Feb. 2016 from Human Phenotype Ontology (HPO) project \cite{Kohler2014}. 3,707 phenotype ontologies at level 8 and their parents, 2,241 phenotype ontologies at level 7, were applied in the following experiments. Two versions of human PPI  networks were obtained from the BIOGRID (Feb. 2016 and Sep. 2016) and two versions of 296 human KEGG pathways (Feb. 2016 and Sep. 2016) were extracted for evaluation. A table summary of the data used in this paper can be found in {\emph{Supporting Information}}.

%The details of data used in our experiments has been summarized in Table  \ref{Tab:Data}.
%\begin{table}[h]
%  \caption{Data Description}
%  \label{Tab:Data}
%  \begin{center}
%  \begin{tabular}{c|c|c|c|c}
%    \hline
%    % after \\: \hline or \cline{col1-col2} \cline{col3-col4} ...
%    Dataset & Relations & Number of  & Number of & Number of \\
%            &(A-B)          &A              &B          &(A-B) \\
%    \hline\hline
%    \multirow{4}*{mouse}
%    & Gene-Phenotype(parent) & 1350 & 2557 & 6889 \\\cline{2-5}
%    & Gene-Phenotype(child) & 1350 & 3414 & 8635 \\\cline{2-5}
%    & Parent-child Phenotype Relation & 2557 & 3414 & 2995 \\\cline{2-5}
%    & PPI (Gene-Gene) (Feb. 2016)& 6234 & 6234 & 14715 \\\cline{2-5}
%    & PPI (Gene-Gene) (Sep. 2016)& 6234 & 6234 & 26190 \\\cline{2-5}
%    & Genes-Pathway (Feb. 2016) & 7753 & 292 & - \\\cline{2-5}
%    & Genes-Pathway (Sep. 2016) & 7882 & 292 & - \\\cline{2-5}
%    %& Gene-Phenotype(PPI) &670 & 5420 & 16122 \\
%    \hline
%    \hline
%    \multirow{4}*{human}
%     & Gene-Phenotype(parent) & 3280 & 2241 & 23532 \\\cline{2-5}
%    & Gene-Phenotype(child) & 3280 & 3707 & 30397 \\\cline{2-5}
%    & Parent-child Phenotype Relation & 2241 & 3707 & 2702 \\\cline{2-5}
%    & PPI (Gene-Gene) (Feb. 2016)& 20585 & 20585 & 261847 \\\cline{2-5}
%    & PPI (Gene-Gene) (Sep. 2016)& 20585 & 20585 & 483838 \\\cline{2-5}
%    & Genes-Pathway (Feb. 2016) & 6989 & 296 & - \\\cline{2-5}
%    & Genes-Pathway (Sep. 2016) & 7087 & 296 & - \\\cline{2-5}
%    %& Gene-Phenotype(PPI) &3193 & 7143 & 333616 \\
%    \hline
%  \end{tabular}
%  \end{center}
%\end{table}
\subsection*{\textbf{Problem Formulation}}
The notations used in the models are summarized in Table \ref{Tab:Notations}. Let $n$ be the number of genes, $m$ be the number of phenotypes, and the gene-phenotype associations are represented by a binary matrix $\bm{A}_{(n \times m)}$ with 1 for entries of known associations and 0 otherwise. The loss function of factorizing matrix $\bm{A}$ is used to derive gene clusters $\bm{G}_{(n \times k)}$ and phenotype clusters $\bm{P}_{(k \times m)}$ based on gene-phenotype associations, where $k$ is the number of clusters. The loss function of factorizing matrix $\bm{A}$ can be defined as:
\begin{equation}\label{equ:NMF}
\begin{split}
L(\bm{P},\bm{G}) =  ||\bm{A}-\bm{GP}||^{2}_{F}.
\end{split}
\end{equation}

However, the loss function mentioned above does not consider the hierarchical structure of phenotype ontologies. To address this problem, we design a framework by considering phenotypes at different levels in the separate hierarchical structure separately (Fig. \ref{fig:model}(a)). In this framework, we assume the gene clustering derived from both parent and child phenotype ontology levels should be identical (Fig. \ref{fig:model}(b)). In addition, a phenotype mapping constraint is also considered to reinforce the consistency between learned phenotype clusters at parent level and those at child level
(Fig. \ref{fig:model}(c), mapping relations are represented by dot lines). By optimizing the components mentioned above, we propose a CMNMF algorithm to learn the gene clusters from gene-phenotype associations with different levels of phenotype ontologies.

\begin{table}[t!]
 \caption{Summary of notations.}\label{Tab:Notations}
%\processtable{Summary of notations.\label{Tab:01}} {
\begin{tabular}{|l|l|}
    \hline
    Notations & Explanations\\
    \hline\hline
    $\bm{A}$ & Genome-phenome association matrix\\
    $\bm{A}_1$ & Genome-phenome association matrix with phenotype\\
    & ontology at parent level\\
    $\bm{A}_2$ & Genome-phenome association matrix with phenotype\\
    & ontology at child level\\
    $\bm{G}$ & Gene cluster membership matrix\\
    $\bm{P}$ & Phenotype cluster membership matrix\\
    $\bm{G}_0$ & Annotated gene cluster membership matrix\\
    $\bm{P}_1$ & Phenotype cluster membership matrix at parent level\\
    $\bm{P}_2$ & Phenotype cluster membership matrix at child level\\
    $\bm{M}$ & Phenotype ontologies relationship matrix\\
    $n$ & Number of genes\\
    $m$ & Number of phenotypes\\
    $k$ & Number of latent clusters\\
    $m_1$ & Number of phenotypes at parent level\\
    $m_2$ & Number of phenotypes at child level\\
    \hline
  \end{tabular}
\end{table}

\subsection*{\textbf{Loss Functions for Penalizing Inconsistency}}
Motivated by the assumption mentioned above, the phenotype ontologies are divided into two sets according to the two adjacent levels. Two gene-phenotype association matrices $(\bm{A}_1)_{n\times m_1}$ and $(\bm{A}_{2})_{n\times m_2}$ are set up based on the original gene-phenotype association matrix $\bm{A}$ and the two sets of phenotype ontologies. We assume that the gene clusters $\bm{G}_{n\times k}$, derived by matrix factorization on $\bm{A}_{1}$ and $\bm{A}_{2}$, should be consistent, although the genes are annotated by phenotype ontologies at adjacent levels. $\bm{G}_{n\times k}$ can be derived by optimizing the following loss function:
\begin{equation}\label{equ:L_O}
{L_C} = \mathop {\min }\limits_{\bm{G},{\bm{P}_1},{\bm{P}_2}} \left\| \bm{A}_1 - \bm{G{P}}_1 \right\|_F^2 + \alpha \left\| {\bm{A}_2} - \bm{GP}_2 \right\|_F^2,
\end{equation}
where $\alpha$ is a hyper-parameter to balance the two matrix factorization problems. To reinforce the hierarchical mapping relationships between phenotypes at parent level and child level, the hierarchical mapping constraint on phenotype ontologies is added to the loss function,
\begin{equation}\label{equ:L_H}
\begin{split}
%{L_H} = \sum\limits_{ij} {{\bm{M}_{ij}}} {(\bm{P}_1^{(i)})^T}\bm{P}_2^{(j)} = tr({\bm{P}_1}\bm{MP}_2^T),\
{L_H} &= \sum\limits_{ij} {{\bm{M}_{ij}}} ||\bm{P}_1^{(i)}-\bm{P}_2^{(j)}||^{2} \\
        %&= \sum_{ij}\bm{M}_{ij}(\bm{P}_1^{(i)})^{T}\bm{P}_1^{(i)}
%        +\sum_{ij}\bm{M}_{ij}(\bm{P}_2^{(j)})^{T}\bm{P}_2^{(j)}
%        -\sum_{ij}2\bm{M}_{ij}(\bm{P}_1^{(i)})^{T}\bm{P}_2^{(j)}\\
%        &=tr(\bm{P}_1\bm{D}_1\bm{P}_1^{T})+tr(\bm{P}_2\bm{D}_2\bm{P}_2^{T})-2tr({\bm{P}_1}\bm{MP}_2^T)
\end{split}
\end{equation}
%where $(\bm{D}_1)_{m_1\times m_1}$ and $(\bm{D}_2)_{m_2\times m_2}$ are diagonal matrices with $(\bm{D}_1)_{ii} = \sum_{j}^{}\bm{M}_{ij}$ and $(\bm{D}_2)_{jj} = \sum_{i}^{}\bm{M}_{ij}$ respectively.
$\bm{M}_{m_1\times m_2}$ denotes the hierarchical mapping relation matrix between phenotype ontologies at adjacent levels. $\bm{M}_{ij}$ is set to 1 if there is a parent-child association between phenotype $i$ and phenotype $j$, otherwise 0. We reinforce the hierarchical mapping constraint by maximizing the similarity between the phenotype ontologies with parent-child mapping relation in gene-phenotype network $\bm{A}_{1}$ and $\bm{A}_{2}$. By combining the two components, the loss function can be formulated as follows:
\begin{equation}\label{equ:R-CMNMF1}
\begin{split}
%{L} =& \left\| \bm{A}_1 - \bm{G{P}}_1 \right\|_F^2 + \alpha \left\| {\bm{A}_2} - \bm{GP}_2 \right\|_F^2-\beta tr({\bm{P}_1}\bm{MP}_2^T)\\
{L} =& \left\| \bm{A}_1 - \bm{G{P}}_1 \right\|_F^2 + \alpha \left\| {\bm{A}_2} - \bm{GP}_2 \right\|_F^2+\beta \sum\limits_{ij} {{\bm{M}_{ij}}} ||\bm{P}_1^{(i)}-\bm{P}_2^{(j)}||^{2}\\
\mathrm{s.t. }\quad
%\quad &\sum_j\bm{G}_{ij}=1,\quad \sum_i{(\bm{P}_1)}_{ij}=1,\quad \sum_i{(\bm{P}_2)}_{ij}=1\\
& \bm{G}\geq 0,\ \bm{P}_1\geq 0,\ \bm{P}_2\geq 0
\end{split}
\end{equation}
where $\beta>0$ is a hyper-parameter to balance the two components.

\subsection*{\textbf{The CMNMF Algorithm}}
To minimize the loss function in Equation (\ref{equ:R-CMNMF1}), an alternative iterative schema is adopted. It solves the problem with respect to one variable while fixing the other variables. In the original NMF \cite{DanielD.Lee2001}, the loss function in Equation (\ref{equ:R-CMNMF1}) is not convex on $\bm{G}$, $\bm{P}_1$, and $\bm{P}_2$ jointly, but it is convex on one variable with the other two fixed.
%but it is convex on $\bm{G}$ with a fixed $\bm{P}_1$ and $\bm{P}_2$, vice versa.
 In the following subsections, the steps of deriving $\bm{G}$, $\bm{P}_1$ and $\bm{P}_2$ are presented separately. The complete CMNMF algorithm is outlined in \textbf{Algorithm \ref{alg:CMNMF}}.

\subsubsection*{\textbf{Computation of $G$ in CMNMF}}
Loss function in Equation (\ref{equ:R-CMNMF1}) can be rewritten as:
\begin{equation}\label{equ:R-CMNMF11}
\begin{split}
{L} =& \left\| \bm{A}_1 - \bm{G{P}}_1 \right\|_F^2 + \alpha \left\| {\bm{A}_2} - \bm{GP}_2 \right\|_F^2+
\beta \big(tr(\bm{P}_1\bm{D}_1\bm{P}_1^{T})+tr(\bm{P}_2\bm{D}_2\bm{P}_2^{T})\\
&-2tr({\bm{P}_1}\bm{MP}_2^T)\big)\\
\end{split}
\end{equation}
where $(\bm{D}_1)_{m_1\times m_1}$ and $(\bm{D}_2)_{m_2\times m_2}$ are diagonal matrices with $(\bm{D}_1)_{ii} = \sum_{j}^{}\bm{M}_{ij}$ and $(\bm{D}_2)_{jj} = \sum_{i}^{}\bm{M}_{ij}$ respectively. When variables $\bm{P}_1$ and $\bm{P}_2$ are fixed, the partial derivative of Equation (\ref{equ:R-CMNMF11}) with respect to $\bm{G}$ is:
\begin{equation}\label{equ:G_gradient}\nonumber
\begin{split}
\frac{\partial{L}}{\partial{\bm{G}}}=
&-2(\bm{A}_1{\bm{P}_1^T} - \bm{G}{\bm{P}_1}{\bm{P}_1^T})-2\alpha(\bm{A}_2{\bm{P}_2^T} - \bm{G}{\bm{P}_2}{\bm{P}_2^T})
%+2\lambda_1\bm{G}
\end{split}
\end{equation}
and the multiplicative update rule is:
\begin{equation}\label{equ:updating_G}\nonumber
\bm{G}_{ij}\leftarrow \bm{G}_{ij}
\frac{(\bm{A}_1\bm{P}_1^T+\alpha \bm{A}_2\bm{P}_2^T)_{ij}}
{(\bm{GP}_1\bm{P}_1^T+\alpha \bm{GP}_2\bm{P}_2^T
%+\lambda_1\bm{G}
)_{ij}}
\end{equation}
%To satisfy the equality constraint, $\bm{G}_{ij}$ is normalized as $\bm{G}_{ij}\leftarrow\frac{\bm{G}_{ij}}{\sum_{j}\bm{G}_{ij}}$.

%%%%%%%%%%%%%%% Computation of $P_1$ and $P_2$ %%%%%%%%%%%%%%%%%%%%%%%%
\subsubsection*{\textbf{Computation of $\bm{P}_1$ and $\bm{P}_2$ in CMNMF}}
When $\bm{G}$ is fixed, the partial derivatives of Equation (\ref{equ:R-CMNMF11}) with respect to $\bm{P}_1$ and $\bm{P}_2$ are:
\begin{equation}\label{equ:P1_gradient}\nonumber
\begin{split}
%&\frac{\partial{L(\bm{P}_1)}}{\partial{\bm{P}_1}}=
%-2(\bm{G}^T\bm{A}_1-{\bm{G}^T\bm{GP}_1})-\beta\bm{P}_2\bm{M}^T\\
%&\frac{\partial{L(\bm{P}_2)}}{\partial{\bm{P}_2}}=
%-2\alpha(\bm{G}^T\bm{A}_2-{\bm{G}^T\bm{GP}_2})-\beta\bm{P}_1\bm{M}
&\frac{\partial{L(\bm{P}_1)}}{\partial{\bm{P}_1}}=
-2(\bm{G}^T\bm{A}_1-{\bm{G}^T\bm{GP}_1})+2\beta(\bm{P}_1\bm{D}_1- \bm{P}_2\bm{M}^T)
\\
&\frac{\partial{L(\bm{P}_2)}}{\partial{\bm{P}_2}}=
-2\alpha(\bm{G}^T\bm{A}_2-{\bm{G}^T\bm{GP}_2})+2\beta(\bm{P}_2\bm{D}_2- \bm{P}_1\bm{M})
\end{split}
\end{equation}
and the multiplicative update rule is ($\bm{P}_2$ is fixed when we calculate $\bm{P}_1$, vice versa):
\begin{equation}\label{updating_P}\nonumber
\begin{split}
%&(\bm{P}_1)_{ij}\leftarrow (\bm{P}_1)_{ij}
%\frac{(\bm{G}^T\bm{A}_1+\frac{1}{2}\beta \bm{P}_2\bm{M}^T)_{ij}}
%{(\bm{G}^T\bm{GP}_1
%)_{ij}}\\
%&(\bm{P}_2)_{ij}\leftarrow (\bm{P}_2)_{ij}
%\frac{(\alpha \bm{G}^T\bm{A}_2+\frac{1}{2}\beta \bm{P}_1\bm{M})_{ij}}
%{(\alpha \bm{G}^T\bm{GP}_2
%)_{ij}}
&(\bm{P}_1)_{ij}\leftarrow (\bm{P}_1)_{ij}
\frac{(\bm{G}^T\bm{A}_1+\beta \bm{P}_2\bm{M}^T)_{ij}}
{(\bm{G}^T\bm{GP}_1+\beta \bm{P}_1\bm{D}_1)_{ij}}\\
&(\bm{P}_2)_{ij}\leftarrow (\bm{P}_2)_{ij}
\frac{(\alpha \bm{G}^T\bm{A}_2+\beta \bm{P}_1\bm{M})_{ij}}
{(\alpha \bm{G}^T\bm{GP}_2+\beta \bm{P}_2\bm{D}_2)_{ij}}
\end{split}
\end{equation}
%To satisfy the equality constraint, we normalize $(\bm{P}_1)_{ij}$ as $(\bm{P}_1)_{ij}\leftarrow \frac{(\bm{P}_1)_{ij}}{\sum_{i}(\bm{P}_1)_{ij}}
%$ and $(\bm{P}_2)_{ij}$ as $(\bm{P}_2)_{ij}\leftarrow \frac{(\bm{P}_2)_{ij}}{\sum_{i}(\bm{P}_2)_{ij}}$.
For the loss function of original NMF $L(\bm{P},\bm{G}) = ||\bm{A}-\bm{GP}||^{2}_{F}$, it is easy to check that if $\bm{G}$ and $\bm{P}$ are the solutions, then $\bm{GD}$, $\bm{D}^{-1}\bm{P}_1$ will also form a solution for any positive diagonal matrix $\bm{D}$. To eliminate this uncertainty, in practice it will be further required that the Euclidean length of each column vector in matrix $\bm{G}$ is 1 \cite{Xu2003,Cai2011}. The matrix $\bm{P}$ will be adjusted accordingly so that $\bm{GP}$ does not change. This can be achieved by:
\begin{equation}\label{}
\begin{split}
\bm{G}_{ik}\leftarrow \frac{\bm{G}_{ik}}{\sqrt{\sum_{i}^{}\bm{G}_{ik}^{2}}},\quad
\bm{P}_{kj}\leftarrow \bm{P}_{kj}\sqrt{\sum_{i}^{}\bm{G}_{ik}^{2}}
\end{split}
\end{equation}
This strategy has been adopted in CMNMF as well. After the multiplicative updating procedure converges, the
Euclidean length of each column vector in matrix $\bm{G}$ is set to 1 and the matrix $\bm{P}_1$ and $\bm{P}_2$ are adjusted with following rules:
\begin{equation}\label{}
\begin{split}
&\bm{G}_{ik}\leftarrow \frac{\bm{G}_{ik}}{\sqrt{\sum_{i}^{}\bm{G}_{ik}^{2}}}\\
&(\bm{P}_1)_{kj}\leftarrow (\bm{P}_1)_{kj}\sqrt{\sum_{i}^{}\bm{G}_{ik}^{2}},\quad
(\bm{P}_2)_{kj}\leftarrow (\bm{P}_2)_{kj}\sqrt{\sum_{i}^{}\bm{G}_{ik}^{2}}
\end{split}
\end{equation}

\begin{algorithm}[t]
\caption{\textbf{CMNMF}}\label{alg:CMNMF}
\renewcommand{\algorithmicrequire}{\textbf{Input:}}
\renewcommand{\algorithmicensure}{\textbf{Output:}}
\label{alg:pf}
\begin{algorithmic}[1]
\REQUIRE
$\bm{A}_1$: gene-phenotype association matrix at parent level\\
\ \ \ \ $\bm{A}_2$: gene-phenotype association matrix at child level\\
\ \ \ \ {$\alpha$, $\beta$}: hyper-parameters
\ENSURE \text{model parameters} ${\bm{G}}, {\bm{P}_1}, {\bm{P}_2}$
%\STATE $P \leftarrow \{(i,j)|q_i=q_j,y_i\succ y_j\}$
\STATE ${\bm{G}},{\bm{P}_1},{\bm{P}_2}$ $\leftarrow$ random values
%\STATE \text{stop$\_$condition:iterate number is more than $N$ or the change of loss function's value is less than $\varepsilon$}
%\STATE $ite = 1$
\REPEAT
    \STATE Update $\bm{G}_{ij}\leftarrow \bm{G}_{ij}\frac{(\bm{A}_1\bm{P}_1^T+\alpha \bm{A}_2\bm{P}_2^T)_{ij}}{(\bm{G}\bm{P}_1\bm{P}_1^T+\alpha \bm{G}\bm{P}_2\bm{P}_2^T %+\lambda_1\bm{G}
    )_{ij}}    $
    \STATE Update $(\bm{P}_1)_{ij}\leftarrow (\bm{P}_1)_{ij}\frac{(\bm{G}^T\bm{A}_1+\beta \bm{P}_2\bm{M}^T)_{ij}}{(\bm{G}^T\bm{GP}_1+\beta \bm{P}_1\bm{D}_1)_{ij}}$,\\
     Update $(\bm{P}_2)_{ij}\leftarrow (\bm{P}_2)_{ij}\frac{(\alpha \bm{G}^T\bm{A}_2+\beta \bm{P}_1\bm{M})_{ij}}{(\alpha \bm{G}^T\bm{GP}_2+\beta \bm{P}_2\bm{D}_2)_{ij}}$
    %\STATE Normalize $(\bm{P}_1)_{ij}\leftarrow \frac{(\bm{P}_1)_{ij}}{\sum_{i}(\bm{P}_1)_{ij}}, \quad
%(\bm{P}_2)_{ij}\leftarrow \frac{(\bm{P}_2)_{ij}}{\sum_{i}(\bm{P}_2)_{ij}}$
\UNTIL convergence
\STATE Normalize $\bm{G}_{ik}\leftarrow \frac{\bm{G}_{ik}}{\sqrt{\sum_{i}^{}\bm{G}_{ik}^{2}}}$
\STATE Normalize $(\bm{P}_1)_{kj}\leftarrow (\bm{P}_1)_{kj}\sqrt{\sum_{i}^{}\bm{G}_{ik}^{2}},\quad
(\bm{P}_2)_{kj}\leftarrow (\bm{P}_2)_{kj}\sqrt{\sum_{i}^{}\bm{G}_{ik}^{2}}$
\RETURN ${\bm{G}},{\bm{P}_1},{\bm{P}_2}$
\end{algorithmic}
\end{algorithm}

\subsubsection*{\textbf{Parameter Tuning}}
\begin{figure}[!h]
  \centering
  \begin{minipage}[b]{\linewidth}
   \includegraphics[width=\linewidth]{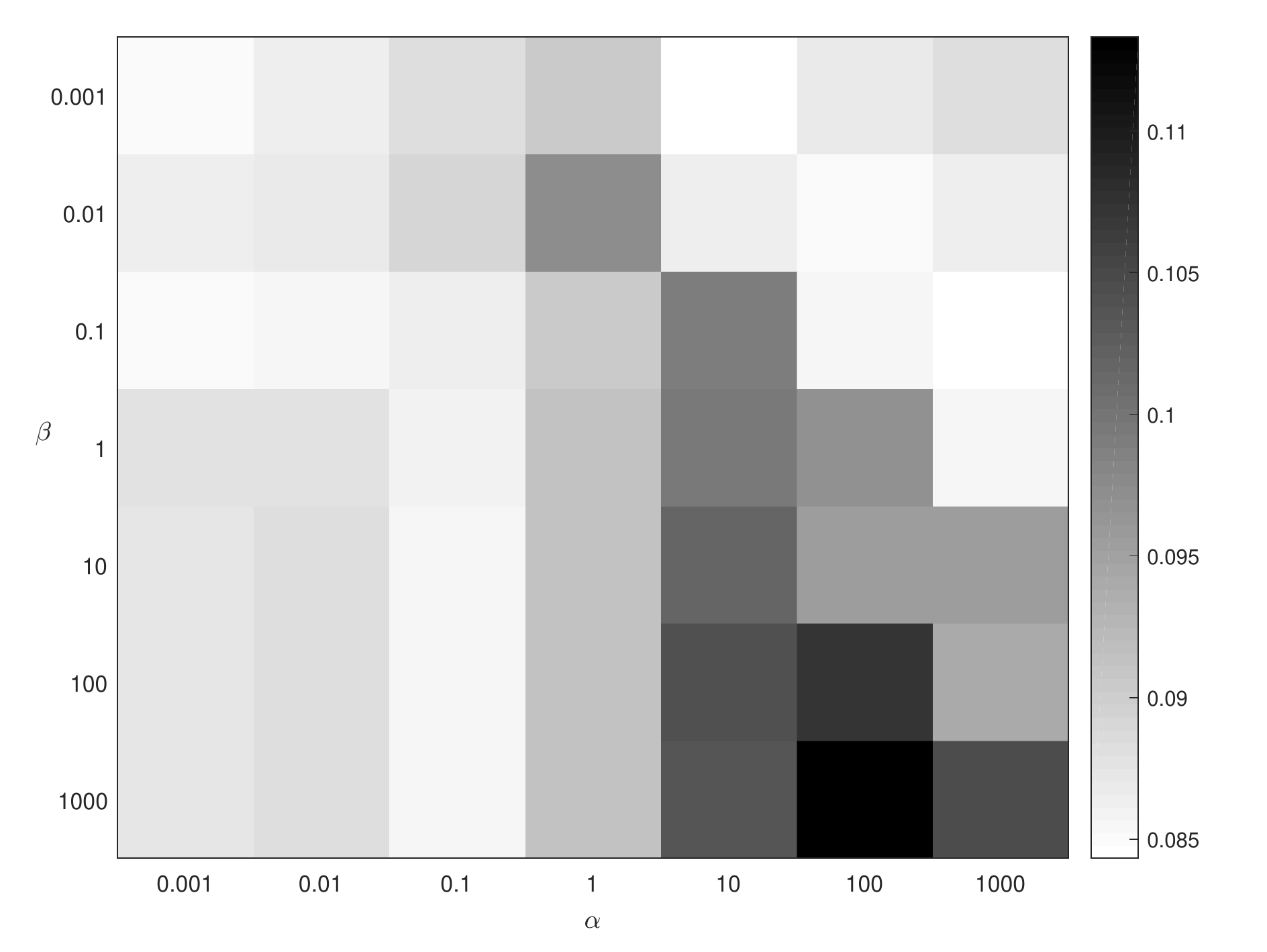}
    %\centerline{ Result based on mouse pathway }
   %\caption{original gene-phenotype association matrix A}
  \end{minipage}
%   \begin{minipage}[b]{.45\linewidth}
%   \includegraphics[width=\linewidth]{mouse_ppi_result.pdf}
%    \centerline{(b) Result based on mouse PPI}
%  \end{minipage}
  \caption{$F_1$ scores under different $\alpha$ and $\beta$ combinations   }
  \label{fig:alpha_beta}
\end{figure}
We use old versions of PPI network (Feb. 2016) and KEGG pathways (Feb. 2016) as validation set to select the best parameters for each method. The selected  parameters have been used in each method to get the performance results by using new versions of PPI network (Sep. 2016) and KEGG pathways (Sep. 2016) as test set.

We perform our experiments on two biological datasets, mouse species dataset and human species dataset. In parameter tuning process, the validation experiment results for each method have been repeated 10 times independently and the average results are applied for parameter tuning. As the parameter tuning processes for mouse data and human data are similar, we show the details on mouse KEGG pathways data as an illustration. Parameter tuning processes on mouse PPI network and human data (human KEGG pathways and human PPI network) are described in {\emph{Supporting Information}}.

The hyper-parameters $\alpha$ and $\beta$ have been tuned by grid with $F_1$ measure. $\alpha$ balances the contributions of two factorization problems on different phenotype ontology levels. When $\alpha$ is close to 0, CMNMF becomes NMF. $\beta$ controls the hierarchical structure effects of phenotype ontologies. When $\beta$ is set to 0, CMNMF becomes ColNMF (Collective NMF) \cite{Singh2008}. The performance of CMNMF with different $\alpha$ and $\beta$ combinations while taking old versions of mouse KEGG pathways (Feb. 2016) as validation set is shown in Fig. \ref{fig:alpha_beta}. We search {$\alpha$ in \{0.001, 0.01, 0.1, 1, 10, 100, 1000\}} and {$\beta$ in \{0.001, 0.01, 0.1, 1, 10, 100, 1000\}}, the darker the color, the higher the $F_1$ score with the corresponding $\alpha$ and $\beta$ combinations. In this experiment, $\alpha=100$ and $\beta = 1000$ are chosen as the best parameters while mouse KEGG pathways data are used as validation set.
%$\alpha=1000$ and $\beta = 0.01$ were chosen while mouse PPI data is used as validation set.
The detailed parameter tuning for baseline methods is described in {\emph{Supporting Information}}.

\subsection*{\textbf{Evaluation}}
There are two types of evaluation indices: external indices and internal indices \cite{Rendon2011}. Because internal indices require extracting additional node features for measuring the similarity between nodes, we choose external indices in the experiments. These external indices, including $F_1$ measure, Jaccard Index, Rand Index, Precision and Recall, are applied to show the consistency between the learned gene clusters and the known KEGG pathways or the gene pairs in the PPI network. The higher the value, the more consistent the learned gene clusters and the known gene sets are.
%Internal indices are used to measure how well the clustering results fit the data without external knowledge, they are based on the information intrinsic to the data exclusively. $\mathcal{M}_{sim}$\cite{Bordino2010}, one of the internal indices, is used for evaluating the results. The higher the $\mathcal{M}_{sim}$, the better the performance. In $\mathcal{M}_{sim}$, genes are labelled with annotated GO terms, and $FS_{simGIC}$ is used to measure the similarity between two genes in the cluster (\emph{{see Supporting Information}}). Genes are well clustered if the labelled GO terms of the genes in a cluster are similar.
\section*{Results}
In this section, we first demonstrate the properties of CMNMF compared with NMF on a small MGI mouse gene-phenotype association matrix. Then CMNMF is compared with seven baseline methods by evaluating the consistency between identified gene modules and the new versions of KEGG pathways (Sep. 2016) or gene pairs in the PPI network (Sep. 2016). Moreover, Gene Ontology enrichment analysis is performed to evaluate the biological significance of discovered gene modules.
%Finally, a supervised CMNMF is proposed for showing the generalization of the model.

\subsection*{\textbf{CMNMF on a Small Mouse Gene-Phenotype Associations}}
\begin{figure}[!h]
  \centering
  \begin{minipage}[b]{.5\linewidth}
   \includegraphics[width=\linewidth]{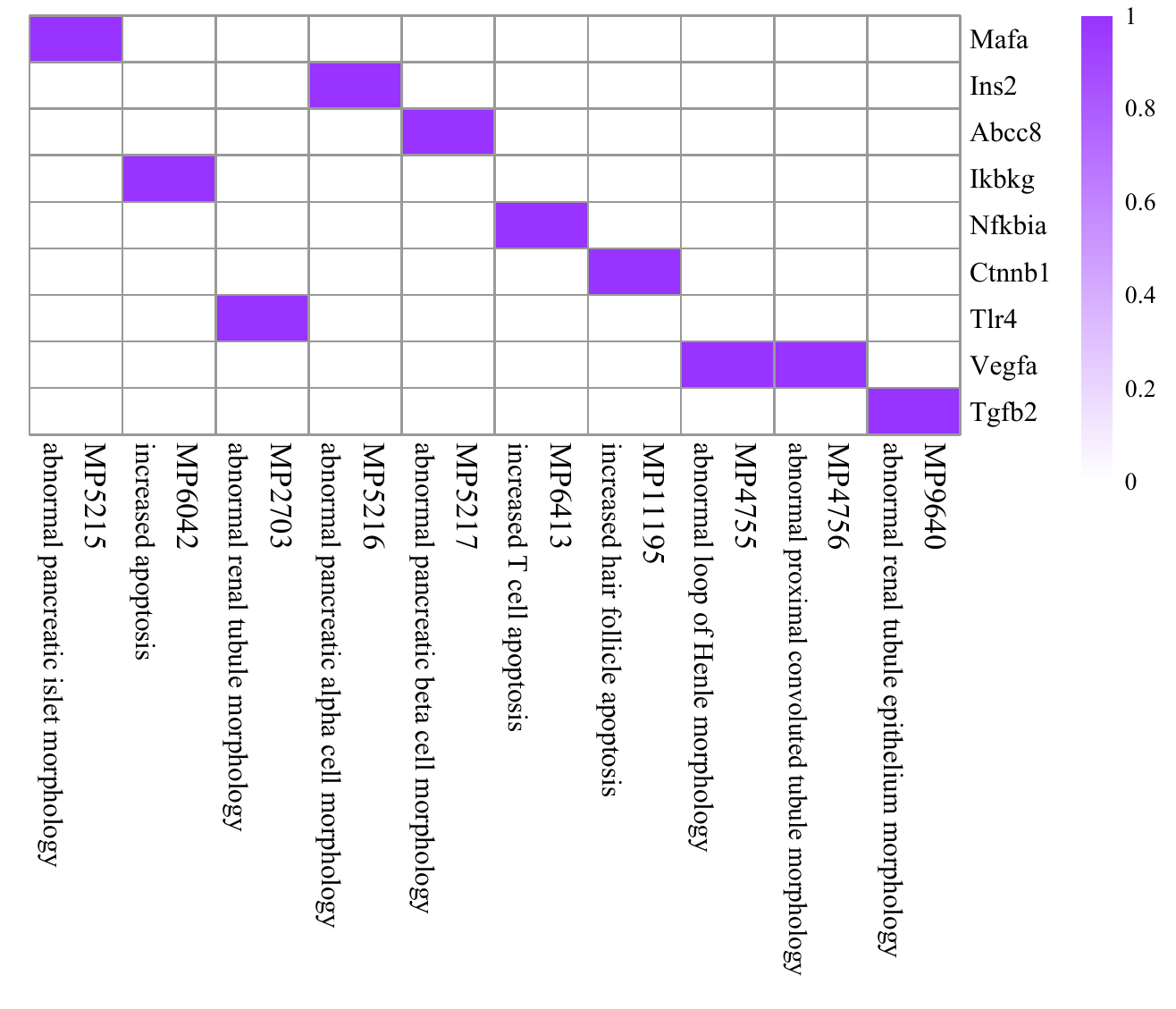}
    \centerline{(a)}
   %\caption{original gene-phenotype association matrix A}
  \end{minipage}
   \begin{minipage}[b]{.45\linewidth}
   \includegraphics[width=\linewidth]{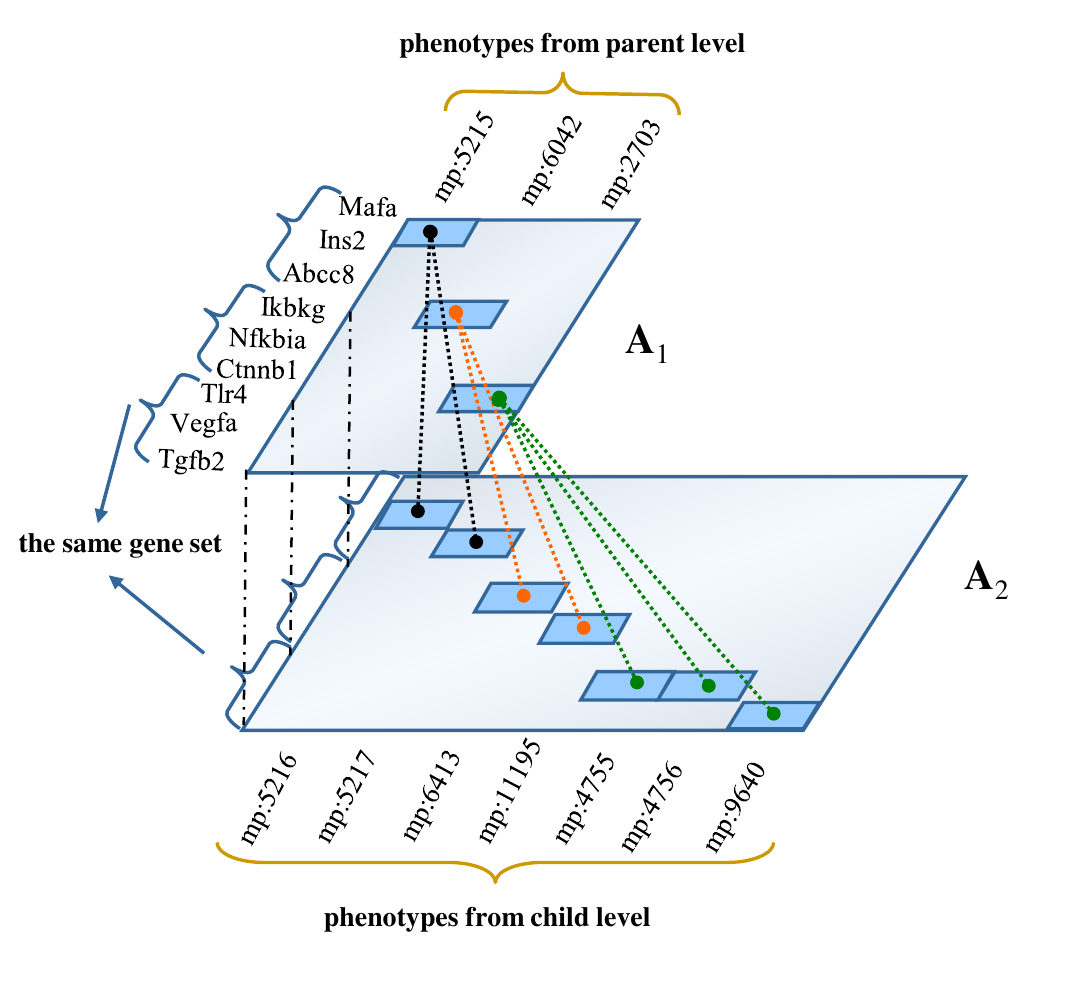}
    \centerline{(b)}
  \end{minipage}
  \hfil
  \begin{minipage}{.3\linewidth}
   \includegraphics[width=\linewidth]{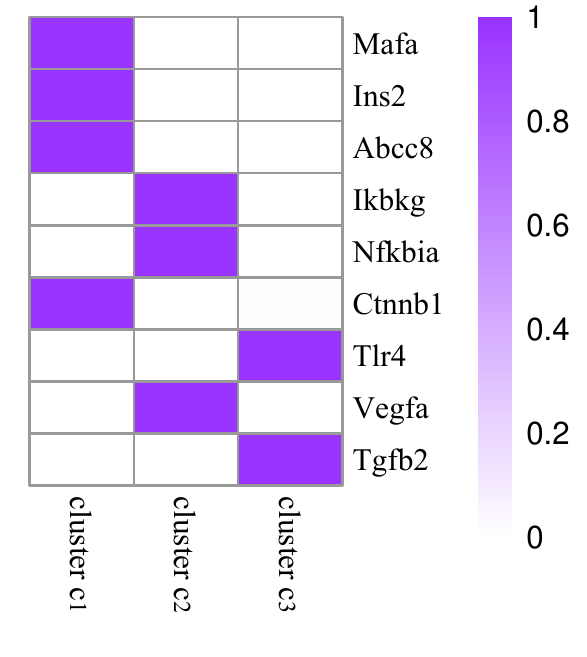}%v20.pdf
   \hfil
    \centerline{(c)}
  \end{minipage}
  \hfil
  \begin{minipage}{.3\linewidth}
   \includegraphics[width=\linewidth]{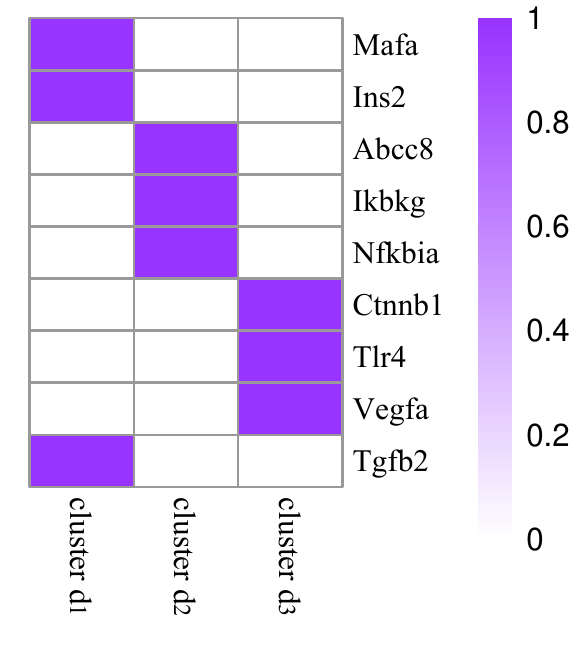}%v23.pdf
    \hfil
    \centerline{(d)}
  \end{minipage}
  \hfil
  \begin{minipage}{.3\linewidth}
   \includegraphics[width=\linewidth]{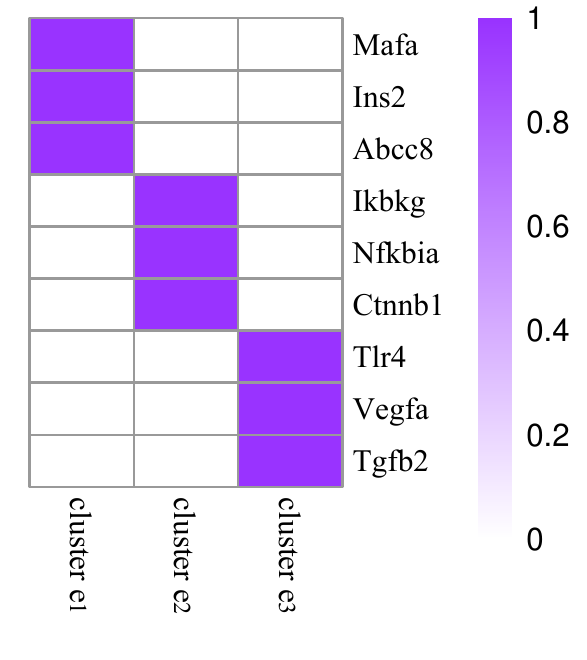}%v24.pdf
    \hfil
    \centerline{(e)}
  \end{minipage}
  \caption{Performance of NMF, CMNMF($\beta$=0) and CMNMF($\beta$=1) on a small \protect \\ gene-phenotype association matrix. (a) Original gene-phenotype association \protect \\ matrix A. (b) Matrix $\bm{A}$ is divided into $\bm{A}_1$ and $\bm{A}_2$ according to the level of  \protect \\phenotype ontology hierarchy. The hierarchical mapping among phenotype \protect \\ ontologies are
   represented by dot lines. (c) Clustering result of NMF.\protect \\ (d) Clustering result of
   CMNMF($\beta$=0). (e) Clustering result of CMNMF($\beta$=1).
   }
  \label{fig:sampled_result}
\end{figure}

To illustrate the effects of consistency constraint (the first two terms in Equation (\ref{equ:R-CMNMF1})) and structure mapping constraint (the last term in Equation (\ref{equ:R-CMNMF1})), we demonstrate the performance of NMF, CMNMF($\beta$=0) and CMNMF($\beta$=1) on a small gene-phenotype association matrix from MGI in Fig. \ref{fig:sampled_result}(a). The gene set in the experiment are selected from three mouse KEGG pathways. In detail, Mafa, Ins2, Abcc8 are from pathway MMU4930 (Type-II diabetes mellitus). Ikbkg, Nfkbia, Ctnnb1 are from MMU5215 (Prostate cancer). Tlr4, Vegfa, Tgfb2 are from MMU5205 (Proteoglycans in cancer). The hierarchical relationships between phenotype ontologies associated with selected genes are shown in Fig. \ref{fig:sampled_result}(b). An effective algorithm should assign the genes from a KEGG pathway into the same cluster.

Fig. \ref{fig:sampled_result}(c), \ref{fig:sampled_result}(d), and \ref{fig:sampled_result}(e) represent the clustering results with NMF, CMNMF($\beta$=0) and CMNMF($\beta$=1), respectively. Compared with Fig. \ref{fig:sampled_result}(c), the significant improvement can be observed by considering the multiple levels of the hierarchy in the phenotype ontology in Fig. \ref{fig:sampled_result}(d) and \ref{fig:sampled_result}(e). By reinforcing the relationship among the phenotypes in different levels, CMNMF($\beta$=1) assigns gene Ctnnb1 to the right cluster comparing with CMNMF($\beta$=0).  Note that the clustering result shown in Fig. \ref{fig:sampled_result}(e) agrees with gene members in the KEGG pathways.

\subsection*{\textbf{Comparison with Baseline Methods by Mining Gene Modules}}
In this section, we evaluate the gene clusters identified by CMNMF with KEGG pathways and PPI network. Seven clustering methods, agglomerative hierarchical clustering (AHC) \cite{Miyamoto1990}, agglomerative hierarchical clustering with pairwise constraints \cite{Miyamoto2010} (Constrained AHC), K-means, pairwise constrained K-means \cite{Basu2004} (Constrained K-means), NMF \cite{Lee1999}, HMF (Hierarchical Matrix Factorization) \cite{AliMashhoori2012} and ColNMF (Collective NMF) \cite{Singh2008}, are compared in the experiment.
Please notice AHC and K-means are unsupervised clustering methods with no parameters,
%other methods may take advantages with more parameters in parameter tuning process.
 in order to have a relatively fair comparison with other methods, we introduce additional pairwise constraints AHC \cite{Miyamoto2010} and pairwise constraints K-means \cite{Wagstaff2001}, the old versions of KEGG pathways (Feb. 2016) and PPI network (Feb. 2016) are used as pairwise constraint validation set to help get clustering results.
%The new versions of KEGG pathways (Sep. 2016) and PPI network (Sep. 2016) are used as test set to get the performance results in Table \ref{tab:measurements_mouse_pathway}-\ref{tab:measurements_human_ppi}.
%To make the results comparable, among all the clustering methods mentioned above, the matrix factorization-based methods are implemented with non-negative constraints.
For CMNMF, HMF and ColNMF, the gene-phenotype association matrix  $\bm{A}$ is divided into two matrices $\bm{A}_1$ and $\bm{A}_2$ according to the levels of phenotype ontologies. For AHC, Constrained AHC, K-means, Constrained K-means, and NMF, the entire gene-phenotype association matrix is applied. Moreover, the associations to parent phenotype terms in the ontology have also been included, i.e. using the ``true path rule" to enrich the association matrix. For NMF, HMF, ColNMF and CMNMF, the gene clustering results are row-normalized by z-score and $G_{ij}$ is set as 0 if it is less than 3.
Six validation indices are reported in Table \ref{tab:measurements_mouse_pathway}-\ref{tab:measurements_human_ppi}.

%%%%%%%%%%%%%%%%%%%%%%%%%%%%%%%%%%%%%%%%%%%%%%%%%%%%%%%%%%%%%%%%%%%%%%%%%%%%
        \begin{table}[!h]
        \centering
        \caption{Evaluation results on mouse KEGG pathways.}\label{tab:measurements_mouse_pathway}
        \begin{tabular}{c|cccccc}
        \hline
        &$F_1$ measure&Precision&Recall&Jaccard Index& Rand Index\\
        \hline
        AHC&0.0871&0.0624&0.1443&0.0455&0.7654\\
        Constrained AHC&0.0925&0.0554&\textbf{0.2803}&0.0485&0.5738\\
        K-means&0.0517&0.0760&0.0392&0.0265&\textbf{0.8886}\\
        Constrained K-means&0.0554&0.0750&0.0439&0.0285&0.8839\\
        NMF&0.1037&\textbf{0.1084}&0.0993&0.0547&0.8668\\
        HMF&0.0818&0.1080&0.0658&0.0426&0.8855\\
        ColNMF&0.0883&0.1074&0.0750&0.0462&0.8799\\
        CMNMF&\textbf{0.1181}&0.0898&0.1726&\textbf{0.0628}&0.8002\\
        % 2, 5, 1, 3, 4
        \hline
        \end{tabular}
        \end{table}
        %%%%%%%%%%%%%%%%%%%%%%%%%%%%%%%%%%%%%%%%%%%%%%%%%%%%%%%%%%%%%%%%%%%%%%%%%%%%
        \begin{table}[!h]
        \centering
        \caption{Evaluation results on mouse PPI.}\label{tab:measurements_mouse_ppi}
        \begin{tabular}{c|cccccc}
        \hline
        &$F_1$ measure&Precision&Recall&Jaccard Index& Rand Index\\
        \hline
        AHC&0.0037&0.0019&0.1420&0.0019&0.8321\\
        Constrained AHC& 0.0080&0.0040&\textbf{0.6491}&{0.0040}&0.6421\\
        K-means&0.0041&0.0022&0.0426&0.0021&0.9544\\
        Constrained K-means&0.0107&0.0056&0.1197&0.0054&0.9509\\
        NMF&0.0126&0.0065&0.1988&0.0063&0.9305\\
        HMF&0.0137&0.0072&0.1298&0.0069&\textbf{0.9584}\\
        ColNMF&0.0120&0.0062&0.1420&0.0060&0.9478\\
        CMNMF&\textbf{0.0138}&\textbf{0.0072}&0.1521&\textbf{0.0070}&0.9518\\
        % 2, 5, 1, 3, 4
        \hline
        \end{tabular}
        \end{table}
\begin{table}[!h]
        \centering
        \caption{Evaluation results on human KEGG pathways.}\label{tab:measurements_human_pathway}
        \begin{tabular}{c|cccccc}
        \hline
        &$F_1$ measure&Precision&Recall&Jaccard Index& Rand Index\\
        \hline
        AHC&0.0808&0.0998&0.0679&0.0421&\textbf{0.9044}\\
        Constrained AHC&0.0952&0.0570&\textbf{0.2892}&0.0500&0.6599\\
        K-means&0.0787&0.0829&0.0748&0.0409&0.8915\\
        Constrained K-means&0.0889&0.0813&0.0980&0.0465&0.8756\\
        NMF&0.0966&0.0891&0.1056&0.0508&0.8778\\
        HMF&0.0863&\textbf{0.1034}&0.0741&0.0451&0.9029\\
        ColNMF&0.0850&0.0869&0.0831&0.0444&0.8893\\
        CMNMF&\textbf{0.1046}&0.0727&0.1886&\textbf{0.0552}&0.8023\\
        % 2, 5, 1, 3, 4
        \hline
        \end{tabular}
        \end{table}
        %%%%%%%%%%%%%%%%%%%%%%%%%%%%%%%%%%%%%%%%%%%%%%%%%%%%%%%%%%%%%%%%%%%%%%%%%%%%
        \begin{table}[!h]
        \centering
        \caption{Evaluation results on human PPI.}\label{tab:measurements_human_ppi}
        \begin{tabular}{c|cccccc}
        \hline
        &$F_1$ measure&Precision&Recall&Jaccard Index& Rand Index\\
        \hline
        AHC&0.0096&0.0055&0.0410&0.0048&0.9709\\
        Constrained AHC&0.0089&0.0045&\textbf{0.4403}&0.0045&0.6637\\
        K-means&0.0095&0.0051&0.0743&0.0048&0.9464\\
        Constrained K-means&0.0117&0.0068&0.0422&0.0059&\textbf{0.9754}\\
        NMF&0.0141&0.0074&0.1457&0.0071&0.9300\\
        HMF&0.0142&0.0077&0.0927&0.0072&0.9557\\
        ColNMF&0.0156&0.0083&0.1303&0.0079&0.9432\\
        CMNMF&\textbf{0.0166}&\textbf{0.0089}&0.1203&\textbf{0.0084}&0.9510\\
        % 2, 5, 1, 3, 4
        \hline
        \end{tabular}
        \end{table}

\subsubsection*{{Validation by Known KEGG Pathways and Protein-Protein Interactions}}
New versions of KEGG pathways (Sep. 2016) and PPI network (Sep. 2016) are applied as known gene relationships to test the performance of gene clustering results. Table \ref{tab:measurements_mouse_pathway} and Table \ref{tab:measurements_mouse_ppi} show the evaluation results on mouse data with KEGG pathways and PPI network, respectively. The evaluation results on human data are reported in Table \ref{tab:measurements_human_pathway} and Table \ref{tab:measurements_human_ppi}. The best results across all the methods are bold. Comparing with the baseline methods, it is clear that CMNMF outperforms other methods on $F_1$ measure, Jaccard Index in all cases for both mouse and human data. It demonstrates the advantage of combining the consistency constraint with two levels of gene-phenotype association information and the structure constraint with parent-child phenotype ontology mapping information. In particular, comparing with the conventional NMF, the performance of CMNMF is improved with the additional knowledge from the consistency constraint. Moreover, the phenotype structure constraint in CMNMF reinforces the learning results following the mapping relation in the phenotype ontology, so CMNMF gets better performance comparing with ColNMF (CMNMF with $\beta=0$). However, AHC works better than other methods with index ``Recall''. We analyse the clustering results of AHC and find a few large-scale gene clusters (with more than four hundred genes), these large gene clusters would result in an increase in index ``Recall'' and a decrease in ``Precision''. The \emph{centroid} criterion is applied in AHC which tends to find the pair of clusters that leads to minimum increase in total inter-cluster Euclidean distances when merging the clusters. Therefore the compact clustering results identified by AHC will benefit the ``Recall" score. We also notice CMNMF does not achieve the best performance on ``Rand Index''. As we know, ``Rand Index'' takes true negative gene pairs into consideration, however, in most cases the experiment results are evaluated on what we have known, i.e. the true positive gene pairs. True negative gene pairs are dominant in the original data involved in the experiments (accounting for 95\%-99\% of all gene pairs), this would lead to a bias comparison between different methods. Overall, the CMNMF outperforms all current clustering methods and the improvement is obvious.

\subsubsection*{{Validation by Latest Protein-Protein Interactions}}%Prediction Emerging Protein-Protein Interactions
CMNMF is also tested on the latest protein-protein interactions added between Feb. 2016 and Sep. 2016 from BIOGRID. The parameters $\alpha$ and $\beta$ are tuned by the old version PPI network (Feb. 2016) mentioned in the previous section with $F_1$ measure. The results are reported in Table \ref{tab:measurements_mouse_latest_ppi} and Table \ref{tab:measurements_human_latest_ppi} for mouse and human data, with the best $\alpha$ and $\beta$ respectively.
%Comparing with baseline methods, CMNMF also outperforms them on $F_1$ measure, Jaccard Index ,Precision and $\mathcal{M}_{sim}$ in human PPI data.
Comparing with baseline methods, CMNMF also outperforms them on $F_1$ measure, Jaccard Index and Precision.
%It should be noticed that the best method in mouse PPI data is constrained AHC. As we have discussed previously, AHC merges clusters according to Ward's minimum variance criterion, which often results in clusters consisting of hundred of genes, such big clusters have limited significance in practice.
\begin{table}[!h]
        \centering
        \caption{Evaluation results on latest mouse PPI.}\label{tab:measurements_mouse_latest_ppi}
        \begin{tabular}{c|cccccc}
        \hline
        &$F_1$ measure&Precision&Recall&Jaccard Index& Rand Index\\
        \hline
        AHC&0.0015&0.0008&0.1373&0.0008&0.8330\\
        Constrained AHC&{0.0034}&{0.0017}&\textbf{0.6569}&{0.0017}&0.6417\\
        K-means&0.0017&0.0009&0.0392&0.0009&0.9577\\
        Constrained K-means&0.0045&0.0023&0.1176&0.0022&0.9519\\
        NMF&0.0052&0.0026&0.1936&0.0026&0.9312\\
        HMF&0.0053&0.0027&0.1176&0.0027&\textbf{0.9593}\\
        ColNMF&0.0047&0.0024&0.1324&0.0024&0.9488\\
        CMNMF&\textbf{0.0055}&\textbf{0.0028}&0.1495&\textbf{0.0028}&0.9507\\
        \hline
        \end{tabular}
    \end{table}
%Evaluation with newly increasing human PPI network:\\
    \begin{table}[!h]
        \centering
        \caption{Evaluation results on latest human PPI.}\label{tab:measurements_human_latest_ppi}
        \begin{tabular}{c|cccccc}
        \hline
        &$F_1$ measure&Precision&Recall&Jaccard Index& Rand Index\\
        \hline
        AHC&0.0043&0.0023&0.0388&0.0021&0.9727\\
        Constrained AHC&0.0039&0.0019&\textbf{0.4313}&0.0019&0.6639\\
        K-means&0.0040&0.0021&0.0759&0.0020&0.9431\\
        Constrained K-means&0.0041&0.0022&0.0309&0.0020&\textbf{0.9771}\\
        NMF&0.0057&0.0029&0.1309&0.0029&0.9313\\
        HMF&0.0056&0.0029&0.0791&0.0028&0.9572\\
        ColNMF&0.0063&0.0032&0.1160&0.0032&0.9446\\
        CMNMF&\textbf{0.0065}&\textbf{0.0034}&0.1033&\textbf{0.0033}&0.9524\\
        \hline
        \end{tabular}
    \end{table}
\subsection*{\textbf{Biological Analysis on Gene Clusters}}
We further study the functional roles of the identified human gene clusters with enrichment analysis against Gene Ontology (GO) using DAVID \cite{Dennis2003}. The enriched GO terms by selected gene clusters are reported in Table \ref{tab:GO_Enrichment}, the P-value and FDR adjusted P-value are also presented. It is clear that gene clusters found by CMNMF are biological functionally relevant.
\begin{table}[!h]
            \centering
            \caption{Enrichment anaylsis on gene clusters mined by CMNMF with Gene Ontology}\label{tab:GO_Enrichment}
            \begin{tabular}{cllll}
            \hline
            No &Gene Cluster& Most Related GO Terms &P-value&{FDR}\\
            \hline
            \hline
            \multirow{3}{*}{239}&
            \multirow{3}{11em}{BCL2, EGFR, FOXO1,\\TGFA, IKBKB,PDGFB, \\PDGFRA, CREB3L3, \emph{etc.}}&
             positive regulation of cell proliferation;&2.2E-06&3.0E-03\\
             &&wound healing;&1.3E-05&1.8E-02\\
             &&protein heterodimerization activity;&7.3E-05&7.4E-02\\
            \hline
            \multirow{3}{*}{173}&
            \multirow{3}{11em}{CLAM1, NOS1, ITPR1,\\ATP2A2, CYCS,GRIN2A, \\CACNA1C, COX6A1, \emph{etc.}}&
             regulation of cardiac muscle contraction &6.8E-07&9.3E-04
             \\&&by regulation of the release of sequestered calciumion;\\
             &&regulation of ryanodine-sensitive calcium-release&6.8E-07&9.3E-04\\&&channel activity;\\
             &&regulation of cardiac muscle contraction;&9.3E-07&1.3E-03\\
            \hline
            \multirow{3}{*}{143}&
            \multirow{3}{11em}{BRAF, SMAD3, NRAS,\\RAF1, SOS1, TGFB2,\\PTPN11, TGFBR1, \emph{etc.}}&
             intracellular;&2.3E-04&2.1E-01\\
             &&MAPK cascade;&2.9E-04&4.0E-01\\
             &&regulation of Rho protein signal transduction;&8.3E-04&1.1E-00\\
            \hline
            \multirow{3}{*}{140}&
            \multirow{3}{11em}{CDK4, ERCC1, G6PC,\\ GAD2,GHR, GPI1,\\ PDX1,MAFA, \emph{etc.}}&
             mitochondrial respiratory chain complex I assembly;&5.7E-33&7.6E-30\\
             &&mitochondrial inner membrane;&5.6E-16&6.0E-13\\
             &&mitochondrion;&1.8E-09&1.9E-06\\
            \hline
            \multirow{3}{*}{127}&
            \multirow{3}{11em}{GRIN1, HTT, LAMC3,\\DCTN1, SOD1, TBP,\\SLC25A4,NDUFA1, \emph{etc.}}&
             extracellular matrix organization;&1.5E-07&2.1E-04\\
             &&cell adhesion;&1.0E-05&1.4E-02\\
             &&basement membrane;&2.0E-05&1.9E-02\\
            \hline
            \multirow{3}{*}{2}&
            \multirow{3}{11em}{CYTB,ND5,ND2,\\COX1,ND1,COX3,\\ND6,ATP6, \emph{etc.}}&
             mitochondrial electron transport, NADH to ubiquinone;&2.1E-11&2.0E-08\\
             &&NADH dehydrogenase (ubiquinone) activity;&2.2E-11&1.5E-08\\
             &&mitochondrial respiratory chain complex I;&5.8E-09&4.3E-06\\
            \hline
            \end{tabular}
            \begin{tablenotes}
      \small
      \item The first column is the index of gene clusters identified by CMNMF. For each gene cluster, a part of genes are presented in the second column. Three most related GO terms, corresponding P-values and false discovery rate (FDR) are reported in the following columns.
    \end{tablenotes}
            \end{table}

By matching the corresponding gene cluster, the genes in No. 239 cluster have been found associated with prostate cancer.
McDonnell found that BCL2 expression was augmented following androgen ablation and was correlated with the progression of prostate cancer from androgen dependence to androgen independence \cite{McDonnell1992}.
Upregulation of epidermal growth factor receptor (EGFR) and subsequent increased in extracellular-regulated kinase (ERK) and Akt signaling were implicated in prostate cancer progression \cite{Gan2010}.
Zhang showed that the FOXO1 protein, a key downstream effector of the tumor suppressor PTEN, inhibited the transcriptional activity of Runx2 in prostate cancer cells \cite{ZhangHaijun2011}.
Liu examined the effects of TGFA and EGF on both cell proliferation and 3H-thymidine incorporation in an androgen-dependent human prostate cancer cell line, ALVA101, in serum-free medium \cite{Liu1993}.

By matching the corresponding gene cluster, the genes in No. 2 cluster have been found associated with Parkinson disease.
Tanaka analyzed amino acid variations in the cytochrome b (CYTB) molecule of 64 Japanese centenarians and found out the most striking feature of centenarian CYTB was the rareness of amino acid variations in contrast to the variety of amino acid replacements in patients with Parkinson's disease \cite{Tanaka2002}.
%The presence or absence of amino acid changing mutations correctly classified 15 of 16 samples.
Heteroplasmic mutations in a specific region of ND5 largely segregated Parkinson's disease from controls and might be of major pathogenic importance in idiopathic Parkinson's disease \cite{Parker2005}.
Ksel found a heteroplasmic mtDNAG5460A missense mutation in the ND2 subunit gene of NADH dehydrogenase was three times more frequent in Parkinson cases (4/21) compared to controls (5/77). In Ksel's study, he provided a hint that the ND25460 mutation, in combination with other factors, could play a role in disease pathogenesis in a subset of Parkinson patients \cite{Ksel1996}.
Teismann used acetylsalicylic acid, a COX-1/COX-2 inhibitor, to study the possible role of the isoenzymes of cyclooxygenase COX-1 and COX-2 in the MPTP  mouse model of Parkinson's disease \cite{Teismann2001}.

%Prostate BCL2 : McDonnell finds that BCL2 expression is augmented following androgen ablation and is correlated with the progression of prostate cancer from androgen dependence to androgen independence. \cite{McDonnell1992}
%Upregulation of epidermal growth factor receptor (EGFR) and subsequent increases in extracellular-regulated kinase (ERK) and Akt signaling are implicated in prostate cancer progression \cite{Gan2010}.
%Prostate FOXO1: Zhang showed that the forkhead box O (FOXO1) protein, a key downstream effector of the tumor suppressor PTEN, inhibits the transcriptional activity of Runx2 in prostate cancer cells \cite{ZhangHaijun2011}.
%Liu examined the effects of 5 alpha-dihydrotestosterone (DHT) and testosterone (T), EGF, and EGFA on cell proliferation and 3H-thymidine incorporation in an androgen-dependent human prostate cancer cell line, ALVA101, in serum-free medium \cite{Liu1993}.

\section*{Conclusions}
Our results indicate that the gene clusters identified by CMNMF are biologically relevant with the additional information from phenotype ontology hierarchical structure. The evaluation experiments on KEGG pathways and PPI network of mouse and human data show the advantage of CMNMF over other baseline methods. The experiment on the latest PPI dataset shows the possibility of identifying new protein-protein interactions. Furthermore, the gene modules identified by CMNMF are enriched with biologically relevant functions.
%At last, a preliminary supervised CMNMF is proposed for showing its generalization ability.

\renewcommand{\enotesize}{\normalsize}
%\theendnotes

%%%%%%%%%%%%%%%%%%%%%%%%%%%%%%%%%%%%%%%%%%%%%%
%%                                          %%
%% Backmatter begins here                   %%
%%                                          %%
%%%%%%%%%%%%%%%%%%%%%%%%%%%%%%%%%%%%%%%%%%%%%%
\section*{Declarations}
\begin{backmatter}

\section*{Acknowledgements}
We would like to thank Rui Kuang, Wei Zhang for their help with this work.
This work is supported by the National Natural Science Foundation of China (No. 61300166 and No. 61105049), the Natural Science Foundation of Tianjin (No. 14JCQNJC00600), and the Science and Technology Planning Project of Tianjin (No. 13ZCZDGX01098).
\section*{Open Access}
This article is distributed under the terms of the Creative Commons Attribution 4.0 International License (http://creativecommons.org/licenses/by/4.0/), which permits unrestricted use, distribution, and reproduction in any medium, provided you give appropriate credit to the original author(s) and the source, provide a link to the Creative Commons license, and indicate if changes were made. The Creative Commons Public Domain Dedication waiver (http://creativecommons.org/publicdomain/zero/1.0/) applies to the data made available in this article, unless otherwise stated.

\section*{Competing interests}
The authors declare that they have no competing interests.

\section*{Author's contributions}
MQX and YGZ originally design the model. YGZ worked on the method, experiment, analyses, and writing of the manuscript. YJX  contributed on method and analyses.  XF, YXH and JHL contributed on the experiment. ZCH contributed on the method. MQX and YLH contributed on writing of the manuscript. All authors read and approved the final manuscript.
\section*{Additional Files}
  \subsection*{Additional file 1 --- Supporting Information (SI)}
    This supplement file includes some additional detailed illustrative text and tables we have mentioned in our paper. We give clearly explanatory text in Supporting Information, please refer to it for more details.
%%%%%%%%%%%%%%%%%%%%%%%%%%%%%%%%%%%%%%%%%%%%%%%%%%%%%%%%%%%%%
%%                  The Bibliography                       %%
%%                                                         %%
%%  Bmc_mathpys.bst  will be used to                       %%
%%  create a .BBL file for submission.                     %%
%%  After submission of the .TEX file,                     %%
%%  you will be prompted to submit your .BBL file.         %%
%%                                                         %%
%%                                                         %%
%%  Note that the displayed Bibliography will not          %%
%%  necessarily be rendered by Latex exactly as specified  %%
%%  in the online Instructions for Authors.                %%
%%                                                         %%
%%%%%%%%%%%%%%%%%%%%%%%%%%%%%%%%%%%%%%%%%%%%%%%%%%%%%%%%%%%%%
% if your bibliography is in bibtex format, use those commands:
\bibliographystyle{bmc-mathphys} % Style BST file (bmc-mathphys, vancouver, spbasic).
\bibliography{bmc_CMNMF}      % Bibliography file (usually '*.bib' )
% for author-year bibliography (bmc-mathphys or spbasic)
% a) write to bib file (bmc-mathphys only)
% @settings{label, options="nameyear"}
% b) uncomment next line
%\nocite{label}
%%                               %%
%% Additional Files              %%

\end{backmatter}

\end{document}